\documentclass[conference]{IEEEtran}
\IEEEoverridecommandlockouts
\usepackage{cite}
\usepackage{amsmath,amssymb,amsfonts}
\usepackage{graphicx}
\usepackage{textcomp}
\usepackage{xcolor}
\usepackage{url}
\usepackage[noend]{algpseudocode}
\usepackage[ruled,vlined,algo2e]{algorithm2e} 
\def\BibTeX{{\rm B\kern-.05em{\sc i\kern-.025em b}\kern-.08em
	T\kern-.1667em\lower.7ex\hbox{E}\kern-.125emX}}
\begin{document}

\title{Towards a General Deep Feature Extractor for Facial Expression Recognition\\
}

\author{\IEEEauthorblockN{Liam Schoneveld}
	\IEEEauthorblockA{\textit{Powder AI Research} \\
		Paris, France \\
		liam@powder.gg}
	\and
	\IEEEauthorblockN{Alice Othmani}
	\IEEEauthorblockA{\textit{Universit\'e Paris-Est, LISSI, UPEC} \\
		Vitry sur Seine, France \\
		alice.othmani@u-pec.fr}
}

\maketitle

\begin{abstract}
	The human face conveys a significant amount  of information. Through facial expressions, the face is able to communicate numerous sentiments without the need for verbalisation. Visual emotion recognition has been extensively studied. Recently several end-to-end trained deep neural networks have been proposed for this task. However, such models often lack generalisation ability across datasets. In this paper, we propose the Deep Facial Expression Vector ExtractoR (DeepFEVER), a new deep learning-based approach that learns a visual feature extractor general enough to be applied to any other facial emotion recognition task or dataset. DeepFEVER outperforms state-of-the-art results on the AffectNet and Google Facial Expression Comparison datasets. DeepFEVER's extracted features also generalise extremely well to other datasets -- even those unseen during training -- namely, the Real-World Affective Faces (RAF) dataset.
\end{abstract}

\begin{IEEEkeywords}
	Facial Expression Recognition, Visual emotion recognition, Deep learning, model generalisation, Knowledge distillation
\end{IEEEkeywords}

\section{Introduction}
Emotional Intelligence in Human-Computer Interaction has attracted increasing attention from researchers in multidisciplinary fields including psychology, computer vision, neuroscience, artificial intelligence, and related disciplines. Humans also tend to interact with computers in a face-to-face manner. As such, human facial expressions present an important opportunity to better link humans and computers. Designing interfaces able to understand human expressions and emotions can thus improve Human-Computer Interaction (HCI) overall.

Recently, deep neural network-based approaches have become more popular for facial expression recognition (FER). These approaches have significantly improved the performance over more traditional approaches to computer vision \cite{LiDeep2018,Othmani2021,minaee2019deep} due to their capacity to automatically learn both low and high level descriptors from facial images without manual feature engineering.

Despite considerable advancements in the field, Facial Expression Recognition (FER) has remained a challenging task. Most existing methods still lack generalisation ability across datasets acquired under different conditions. The majority of modern approaches in the emotion recognition literature either learn or fine-tune an end-to-end network that can only be used for a specific dataset, and/or use more general features as input to a more basic model.

To overcome this limitation, it is proposed in this paper to learn an independent feature extractor for images of faces, specialised for facial expressions, that could be employed for any downstream FER task or dataset. This approach achieves its generalisation ability by training on multiple labeled FER datasets, and by employing knowledge distillation (specifically, \emph{self-distillation}), alongside additional unlabeled data for FER. The proposed visual facial expression embedding network is described in Section~\ref{sec:proposed_approach}.  Section~\ref{sec:experiments_results} describes the experimental setting and discusses the results obtained on three FER datasets. Finally, Section~\ref{sec:conclusion} concludes the paper and discusses future work.

\section{Proposed Approach}
\label{sec:proposed_approach}
This paper proposes a general and independent deep facial expression feature extractor called  Deep Facial Expression Vector ExtractoR (DeepFEVER). DeepFEVER is a convolutional neural network (CNN) trained using: (1) two labeled FER datasets, (2) an additional unlabeled dataset and (3) the technique known as \textit{knowledge distillation} \cite{hinton2015}.

Not only does DeepFEVER outperform state-of-the-art results on both the datasets it was trained on, it also generalises very well; DeepFEVER achieves highly competitive results on an additional FER dataset, unseen during training, \textbf{without any fine-tuning}.

DeepFEVER is trained using \textit{knowledge distillation}: a two step process whereby a \textit{teacher} network is trained on the task of interest, and then a (typically smaller) \textit{student} network is trained on predictions made by the teacher. Specifically in this work, we leverage the benefits of \textit{self-distillation}, wherein the student network has the same size as (or at least, is not smaller than) the teacher network. Self-distillation is used in this work to improve the performance of the facial expression embedding network, as well as to provide more flexibility over the final model architecture.

The training procedure for DeepFEVER consists of two phases: the teacher model training phase and the student model training phase. In the teacher model training phase, a FaceNet \cite{Schroff2015} is retrained simultaneously on two different visual FER datasets (Section \ref{sec:teacher}). In the student model training phase, we again train on these two datasets (Section \ref{sec:student}). However, the starting point for the student network can be any CNN architecture (we opt for a DenseNet \cite{DenseNet}). The student is also trained on an additional unlabeled dataset, whose training `labels' are provided by the teacher network.

The output of DeepFEVER is a compact vector of dimension $D_{\text{face}}$. This was set to $D_{\text{face}}=256$ as such generally performed better in experiments. During training this vector serves as a shared input for various output heads that each produce predictions for each of the training tasks. After training, this vector can be extracted from face images to provide features for any downstream FER task. More details about the teacher and the student networks follow.

\subsection{The teacher network}
\label{sec:teacher}

The proposed teacher network starts from a pre-trained FaceNet \cite{Schroff2015}. As FaceNet is for person re-identification, the teacher network is further trained using two more datasets to learn specialised features for FER. These datasets are AffectNet \cite{AffectNet} and Google Facial Expression Comparison (FEC) \cite{GoogleFEC}. Further details about the datasets are provided in Section~\ref{sec:datasets}.

\begin{figure*}
	\includegraphics[width=.97\textwidth]{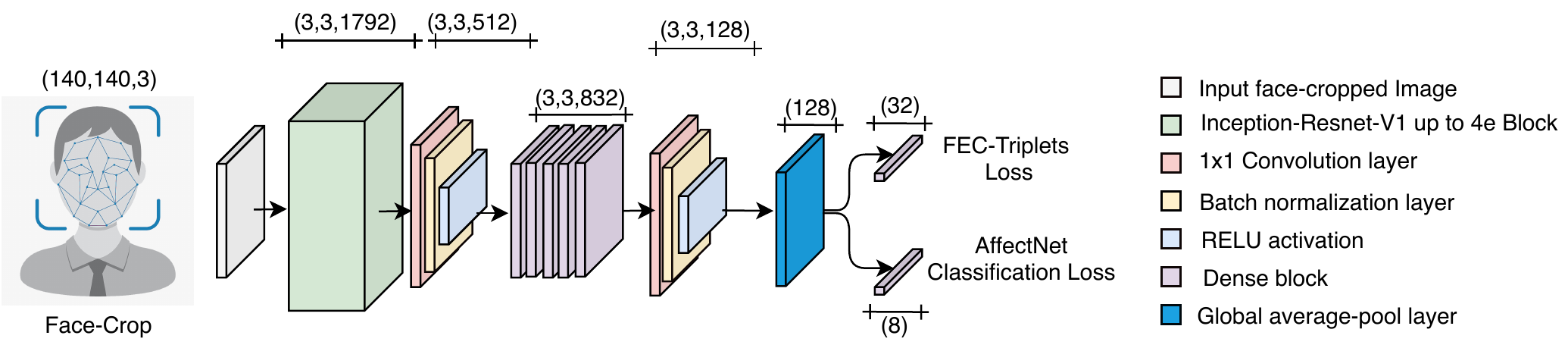}
	\caption{The facial expression recognition neural network architecture, before distillation (i.e. the \emph{teacher network}). Faces are detected and cropped using MTCNN\cite{MTCNN}. The resulting 140x140 RGB images are then fed to an Inception Resnet V1 until the Inception 4e block. This is followed by a 1x1 convolution layer (1x1 Conv), batch normalisation (BN) and a ReLU activation. Then, five DenseNet blocks are applied. Finally, another set of 1x1 Conv, BN and ReLU are applied. The output is then averaged over the spatial dimensions, resulting in a vector of size $D_{\text{face}}$ (in the figure, $D_{\text{face}}=128$). Two separate linear layers then give the final model outputs -- a vector for the Google FEC triplets task, and class logits for AffectNet. The model is trained to minimize both the AffectNet and Google FEC losses simultaneously. The numbers over each block represent the output shape of the tensor after applying that block.}
	\label{fig:fec_archi}
\end{figure*}

The teacher model's architecture (Fig. \ref{fig:fec_archi}) is almost identical to the model proposed in \cite{GoogleFEC}, the only difference is that an additional output head is added for the AffectNet loss. Thus, following \cite{GoogleFEC}, a pre-trained FaceNet\footnote{In this, we used a FaceNet pre-trained on the VGGFace2 dataset, as we found performance to be slightly improved. The pre-trained FaceNet model architecture and weights were obtained from https://github.com/timesler/facenet-pytorch.} is taken up until the Inception 4e block. This is followed by a 1x1 convolution and a series of five untrained DenseNet \cite{DenseNet} blocks. After this, another 1x1 convolution followed by global average pooling reduces this representation to a single $D_{\text{face}}$ dimensional vector. After pooling, two independent linear transformations serve as output heads. These heads take the $D_{\text{face}}$-dimensional facial expression representation vector as input and make separate predictions for the AffectNet and FEC tasks. A 32-dimensional embedding is used for the FEC triplets task, while an 8-dimensional output head produces class logits for AffectNet (which has 8 classes). The teacher network training procedure is detailed in Algorithm \ref{algo1}, while implementation details are provided in Section~\ref{sec:implementation_details}.

To improve the regularisation effects of self-distillation through model ensembling, two teacher networks are trained, and their outputs are concatenated to serve as distillation targets (see Section \ref{sec:student} for details). The only difference between the two teachers are the random seeds used for initialisation, and penultimate layer dimensionalities: one teacher network was trained with $D_{\text{face}}=256$, the other with $D_{\text{face}}=128$.

\begin{algorithm2e}
	
	\SetAlgoLined
	\For{iteration in range($N$)}{
		$(\mathbf{X}_{\text{FEC}}, \mathbf{y}_{\text{FEC}})$ $\leftarrow$ batch of Google FEC triplets and labels \\
		$(\mathbf{X}_{\text{Aff}}, \mathbf{y}_{\text{Aff}})$ $\leftarrow$ batch of AffectNet images and class labels\\
		$\mathbf{e}_{\text{FEC}} \leftarrow f_{\Theta}(\mathbf{X}_{\text{FEC}})$ \Comment{Face embeddings for FEC images}\\
		$\mathbf{e}_{\text{Aff}} \leftarrow f_{\Theta}(\mathbf{X}_{\text{Aff}})$ \Comment{Face embeddings for AffectNet images}\\
		$\mathbf{v}_{\text{FEC}} \leftarrow g_{\phi}(\mathbf{e}_{\text{FEC}})$ \Comment{Predict vectors for triplet loss}\\
		$\mathbf{p}_{\text{Aff}} \leftarrow h_{\theta}(\mathbf{e}_{\text{Aff}})$ \Comment{Predict class probabilities for AffectNet}\\
		$L_\text{FEC} = \texttt{triplet\_loss}(\mathbf{v}_{\text{FEC}},  \mathbf{y}_{\text{FEC}})$\\
		$L_\text{Aff} = \texttt{cross\_entropy\_loss}(\mathbf{p}_{\text{Aff}}, \mathbf{y}_\text{Aff})$\\
		$L = L_\text{FEC} + \alpha * L_\text{Aff}$ \Comment{Total loss for training step}\\
		Obtain all gradients $\Delta_\text{all} = (\frac{\partial L}{\partial \Theta}, \frac{\partial L}{\partial \phi}, \frac{\partial L}{\partial \theta})$\\
		$(\Theta, \phi, \theta) \leftarrow \texttt{SGD}(\Delta_\text{all}$) \Comment{Update feature extractor and output heads' parameters simultaneously}
	}
	\caption{DeepFEVER: training the teacher network. Given feature extractor network $f_{\Theta}$, Google FEC output head $g_{\phi}$, AffectNet output head $h_{\theta}$,  number of training steps $N$, AffectNet loss weight $\alpha$.}
	\label{algo1}
\end{algorithm2e}

\subsection{Student network}
\label{sec:student}

The student network is a DenseNet201 pre-trained on Imagenet.\footnote{We use the DenseNet implementation and pre-trained Imagenet weights provided in the \texttt{torchvision} Python package.} The student network training procedure is essentially the same as described in Algorithm \ref{algo1}, except that we additionally sample batches of unlabeled data from an internal dataset, which we refer to as \textit{PowderFaces} (See Section~\ref{sec:datasets}). The sampled batches of face crops from the Google FEC, AffectNet, and PowderFaces datasets are passed through the two teacher networks. Each of the two teacher networks produces predictions for the Google FEC task (32-dimensional) and AffectNet class logits (8-dimensional). These four vectors (i.e., two vectors from two teacher networks) are individually L2-normalised. The four normalised vectors are then concatenated, producing one long vector of dimension 80. A knowledge distillation loss (we specifically use \textit{Relational Knowledge Distillation} \cite{park2019relational} for the loss function) is then calculated by comparing the output of a third output head in the student network to this 80-dimensional target vector. This knowledge distillation loss is then added to the standard AffectNet and Google FEC losses, which are calculated as per the teacher network training procedure. Implementation details for the student network are provided in Section~\ref{sec:implementation_details}.

\section{Experiments and Results}
\label{sec:experiments_results}

\subsection{Datasets}
\label{sec:datasets}

Two datasets are used to train the teacher network:
\begin{itemize}
	\item \textbf{AffectNet} \cite{AffectNet}, which consists of around 440,000 in-the-wild face crop images, each of which is human-annotated into one of eight facial expression categories (Neutral, Happy, Sad, Surprise, Fear, Disgust, Anger and Contempt).
	\item \textbf{Google Facial Expression Comparison (FEC)} \cite{GoogleFEC}, which consists of around 700,000 triplets of unique face crop images. Annotations denoting the most similar pair of face expressions in each triplet are provided. The goal is to train a model that places the similar pair closer together in a learned embedding space.
\end{itemize}
An additional, unlabeled dataset is used to train the student network via knowledge distillation: the \textbf{PowderFaces dataset}. This dataset was created by downloading approximately 20,000 short, publicly-available videos from various online sources such as YouTube. To increase the frequency of faces in the dataset, specific search terms and topics were used when searching for videos, such as `podcast', `interview', or `monologue'. MTCNN face detection \cite{MTCNN} was then applied to the extracted frames from those videos, producing approximately 1 million individual face crops.

To evaluate the generalisation ability of DeepFEVER, the \textbf{Real-World Affective Faces Database (RAF)} \cite{RAFDataset} is used. RAF consists of 12,271 training face images and 3,068 validation face images, each annotated into one of seven classes (Surprise, Fear, Disgust, Happiness, Sadness, Anger and Neutral).

\subsection{Implementation details}
\label{sec:implementation_details}
\textbf{Implementation details of the teacher network:} The entire network is retrained (i.e., none of the original FaceNet model's weights are frozen). The SGD optimizer is used with a learning rate (LR) of $0.005$ and Nesterov momentum of $0.9$. No learning rate scheduling is used -- rather training was simply stopped after 13 epochs, based on validation performance~\footnote{Here an epoch refers to a full pass over the FEC triplets dataset. To obtain AffectNet samples, we simply continually loop over the AffectNet dataset in the background, reshuffling the dataset at the end of each background loop.}. A batch size of 64 for both the triplets and AffectNet datasets is used. To bring the AffectNet cross-entropy loss' magnitude more in line with that of the triplet loss, it is multiplied by a factor of $\alpha = 0.1$ during training. Dropout of $0.1$ is applied to the $D_{\text{face}}$-dimensional expression vector before passing it to the output heads.

\textbf{Implementation details of the student network:} For the knowledge distillation loss, Relational Knowledge Distillation (RKD) (Park et al.
[2019]) is used. The RKD distance-wise loss is multiplied by 25, and the angle-wise loss is multiplied by 50, to bring their magnitude up to a similar level as per the AffectNet and Google FEC loss components. The implementation details of the student network are exactly the same as per the teacher network (Section~\ref{sec:teacher}), except that it was trained for longer (18 epochs), uses 
smaller batch sizes (36 Google FEC triplets, 16 AffectNet images and 16 PowderFaces images per iteration) and has a penultimate layer dimension of $D_{\text{face}}=256$ (like one of the two teacher networks), with dropout of $0.2$ instead of $0.1$

\textbf{Implementation details of the linear evaluation on RAF:} To evaluate DeepFEVER on RAF, the $D_{\text{face}}$-dimensional feature vectors are extracted using the student model for every face in the RAF dataset. A logistic regression model \cite{sklearn} is then trained (with regularisation parameter $C$ set to $10000$) on the features extracted from the training set. The accuracy of this model is then calculated by predicting the validation set labels, using only the extracted feature vectors as input.

\subsection{Performance of DeepFEVER for visual FER}

We evaluate DeepFEVER on the standard evaluation subsets of the two datasets (AffectNet and Google FEC) used in training, plus a third dataset, unseen during training (RAF):
\begin{enumerate}
	\item \textbf{AffectNet:} for AffectNet, which requires classifying faces into eight discrete facial expression classes, a logistic regression model is trained on the features extracted by the student network for the entire AffectNet training set.\footnote{When training this logistic regression, we re-weight the classes in the AffectNet training set to have equal representation, as per the validation set.} For eight-class classification, DeepFEVER achieves state-of-the-art results on the AffectNet validation set, with an accuracy of $61.6\%$ (Table \ref{tab:affectnet_results}). Some papers report seven-class accuracy by excluding images with the \textit{contempt} class from the training and validation sets. DeepFEVER also achieves state-of-the-art results on this seven-class task, with an accuracy of 65.4\%.
	\item \textbf{Google FEC:} following \cite{GoogleFEC}, models are evaluated using triplet accuracy on the Google FEC test set. As shown in Table~ \ref{tab:fec_results}, DeepFEVER outperforms state-of-the-art methods, with an accuracy of $86.5\%$ .
	\item \textbf{RAF:} A logistic regression trained on DeepFEVER face vectors outperforms all but one of the previous state-of-the-art results on RAF, including those which train directly on the RAF dataset (Table \ref{tab:raf_results}).
\end{enumerate}

\begin{table}[h] 
	\caption{Accuracy of DeepFEVER on the AffectNet validation set (7 or 8 classes) compared to existing state-of-the-art methods}
	\begin{tabular*}{\columnwidth}{l|l|l} 
		\hline 
		\textbf{Methods} & \textbf{8-class} & \textbf{7-class}\\
		\hline
		CNNs and BOVW + local SVM \cite{Georgescu2019}& 59.6\%\\
		
		
		Pyramid with Super Resolution \cite{Vo2020}& 60.7\% & 63.8\%\\
		
		PAENet \cite{Hung2019}&  & 65.3\%\\
		
		DeepFEVER (Teacher model) & 61.3\% & 65.4\%\\
		
		DeepFEVER (Student, no distillation) & 58.8\% & 62.6\%\\
		
		DeepFEVER (Distilled student, no PowderFaces) & 61.1\% & 65.2\%\\
		
		DeepFEVER (Distilled student) &  \textbf{61.6\%} & \textbf{65.4}\%\\
		\hline
	\end{tabular*}
	\label{tab:affectnet_results}
\end{table}

\begin{table}[h] 
	\caption{Triplet accuracy of DeepFEVER on the Google FEC test set compared to the existing state-of-the-art approach}
	\begin{tabular*}{\columnwidth}{l|l} 
		\hline
		\textbf{Methods} & \textbf{Accuracy} \\
		\hline
		Fine-tuned FaceNet \cite{GoogleFEC}& 81.8\% \\
		DeepFEVER (Teacher model) & 84.5\% \\
		DeepFEVER (Student, no distillation) & 85.0\% \\
		DeepFEVER (Distilled student, no PowderFaces) & 86.4\% \\
		DeepFEVER (Distilled student) & \textbf{86.5\%}\\
		\hline
	\end{tabular*}
	
	\label{tab:fec_results}
\end{table}

\begin{table}[h] 
	\caption{Performances of DeepFEVER on the RAF `basic' test set (7 classes) compared to existing state-of-the-art methods}
	\begin{tabular*}{\columnwidth}{l|l} 
		\hline 
		\textbf{Methods} & \textbf{Accuracy}\\
		\hline
		
		IPA2LT \cite{Zeng2018}& 86.8\%\\
		
		RAN-ResNet18 \cite{Wang2020}& 86.9\%\\
		
		PSN \cite{Vo2020}& \textbf{89.0}\%\\
		
		DeepFEVER (Distilled student, \textbf{no fine-tuning}) &  87.4\%\\
		\hline
	\end{tabular*}
	\label{tab:raf_results}
\end{table}

\subsection{Ablation study}

To experimentally verify the importance of the different components of our approach, an ablation study is performed (Tables \ref{tab:affectnet_results} and \ref{tab:fec_results}). Training the student model architecture without the distillation loss component revealed the importance of distillation: without such, performance dropped substantially: to 58.8\% on AffectNet (eight-class) and 85\% on Google FEC.

Similarly, to determine the importance of the unlabeled PowderFaces dataset, the student model was trained with distillation, but without the additional distillation targets provided by using this unlabeled data. The results suggest that the additional unlabeled data is not so important: accuracy on AffectNet dropped only slightly to 61.1\%, and performance on Google FEC reduced by only 0.1\% to 86.4\%. Other works successfully applying similar approaches to self-distillation used an unlabeled dataset on the order of 300 times larger than the labeled dataset \cite{xie2020self}. In our case the ratio of unlabeled data to labeled data is closer to one to one. We leave determining whether larger volumes of unlabeled data can further improve DeepFEVER's performance to future work.


\section{Conclusion and Future work}
\label{sec:conclusion}
This research proposes a new approach to learning a deep facial expression feature extractor, with high generalisation ability across datasets. The proposed DeepFEVER network outperforms state-of-the-art approaches on the datasets it was trained on, and achieves highly competitive results the RAF dataset without any fine-tuning. In future work, we plan to explore scaling up the unlabeled dataset for self-distillation and quantisation of the learned facial embedding space.



\section*{Acknowledgment}

This work was supported by Powder, a Paris-based start-up creating the social network for gamers. \url{https://powder.gg/}

\end{document}